\documentclass[letterpaper, 10 pt, conference]{ieeeconf}  

\IEEEoverridecommandlockouts                              

\usepackage{xcolor}
\usepackage{gensymb}
\usepackage{graphicx} 
\usepackage{amssymb}
\usepackage{booktabs}
\usepackage{tabularx}
\usepackage{multirow}

\title{\LARGE \bf
\textit{dARt Vinci}: Egocentric Data Collection for \\
Surgical Robot Learning at Scale
}

\author{Yihao Liu$^{1,2,*}$, Yu-Chun Ku$^{1,*}$, Jiaming Zhang$^{1,*}$, Hao Ding$^{1}$, \\Peter Kazanzides$^{1}$, Mehran Armand$^{1,2}$
\thanks{This work was supported by National Institute of Arthritis and Musculoskeletal and Skin Diseases (R01AR080315) and National Institute of Biomedical Imaging and Bioengineering (R01EB023939). }
\thanks{$^{1}$ Department of Computer Science, Johns Hopkins University, Baltimore, MD 21218, USA}%
\thanks{$^{2}$ The Institute for Integrative and Innovative Research, University of Arkansas, Fayetteville, AR 72701, USA}
\thanks{Corresponding author {\tt\small (yliu333@jhu.edu)}}
\thanks{* Equal contribution}
}

\begin{document}

\maketitle
\thispagestyle{empty}
\pagestyle{empty}

\begin{abstract}

Data scarcity has long been an issue in the robot learning community. Particularly, in safety-critical domains like surgical applications, obtaining high-quality data can be especially difficult. It poses challenges to researchers seeking to exploit recent advancements in reinforcement learning and imitation learning, which have greatly improved generalizability and enabled robots to conduct tasks autonomously. We introduce \textit{dARt Vinci}, a scalable data collection platform for robot learning in surgical settings. The system uses Augmented Reality (AR) hand tracking and a high-fidelity physics engine to capture subtle maneuvers in primitive surgical tasks: By eliminating the need for a physical robot setup and providing flexibility in terms of time, space, and hardware resources-such as multiview sensors and actuators-specialized simulation is a viable alternative. At the same time, AR allows the robot data collection to be more egocentric, supported by its body tracking and content overlaying capabilities. Our user study confirms the proposed system's efficiency and usability, where we use widely-used primitive tasks for training teleoperation with da Vinci surgical robots. Data throughput improves across all tasks compared to real robot settings by 41\% on average. The total experiment time is reduced by an average of 10\%. The temporal demand in the task load survey is improved. These gains are statistically significant. Additionally, the collected data is over 400 times smaller in size, requiring far less storage while achieving double the frequency.

\end{abstract}

\section{Introduction}

Robotic Minimally Invasive Surgery (RMIS) has become a cornerstone of modern surgery over the past two decades \cite{taylor2016medical, taylor2008medical, ma2021active}. It offers promising potentials in reduced patient recovery times, improved precision, and decreased risk of complications \cite{ma2021active, sefati2020surgical}. There has been growing interest in extending the capabilities of RMIS by automating specific subtasks or even entire surgical procedures \cite{ahmidi2017dataset, hwang2022automating, yu2024orbit, liu2024roadmap}. Such automation may hold the promise of reducing prohibitively expensive medical costs and reducing the cognitive and physical demands on surgeons while improving surgical outcomes. Advancements in robotics, particularly in training robots or using various foundation models to perform manipulation tasks, have demonstrated significant potential for automating actions in complex scenarios \cite{kroemer2021review, wang2023gensim, hua2024gensim2, akkaya2019solving, wu2023tidybot, xiao2023robot}. Among the promising approaches, imitation learning and reinforcement learning stand out as powerful tools for enabling robots to learn and adapt to dynamic tasks \cite{ravichandar2020recent, zhao2023learning}, including those encountered in surgical settings \cite{li2023imitation, kim2024surgical}. 

However, robot learning relies on a large amount of high-quality collected data. In the 1980s, Behavioral Cloning (BC) was developed \cite{pomerleau1988alvinn} and is still one of the simplest imitation learning algorithms. It works by treating imitation learning as a supervised learning problem, and the goal is to map observations to corresponding actions. The groundbreaking study focuses on autonomous driving and uses road snapshots as driving signals. As claimed by the study's researchers, they had already encountered data collection issues as the process was limited by the variety of road conditions and changes in equipment parameters \cite{pomerleau1988alvinn}. Scaling the imitation learning algorithms with more data \cite{brohan2022rt, jang2022bc} and tasks \cite{dasari2021transformers} and consistent algorithmic innovations \cite{mandlekar2021matters, florence2022implicit, rahmatizadeh2018vision} have led to impressive systems that can generalize to new objects, instructions, or scenes. However, high-quality datasets do not come easily. The success of foundation models across different domains shows the critical role of scaling datasets in enhancing model performance and generalization. Yet, RMIS encounters substantial challenges in building large-scale datasets comparable to large datasets like ImageNet. Privacy regulations, institutional policies, and funding constraints are all factors that impede data sharing. Furthermore, surgical robots are typically operated by highly skilled surgeons, further limiting the availability of diverse and extensive data.

\begin{figure*}[ht]
    \centering
    \includegraphics[width=0.9\textwidth]{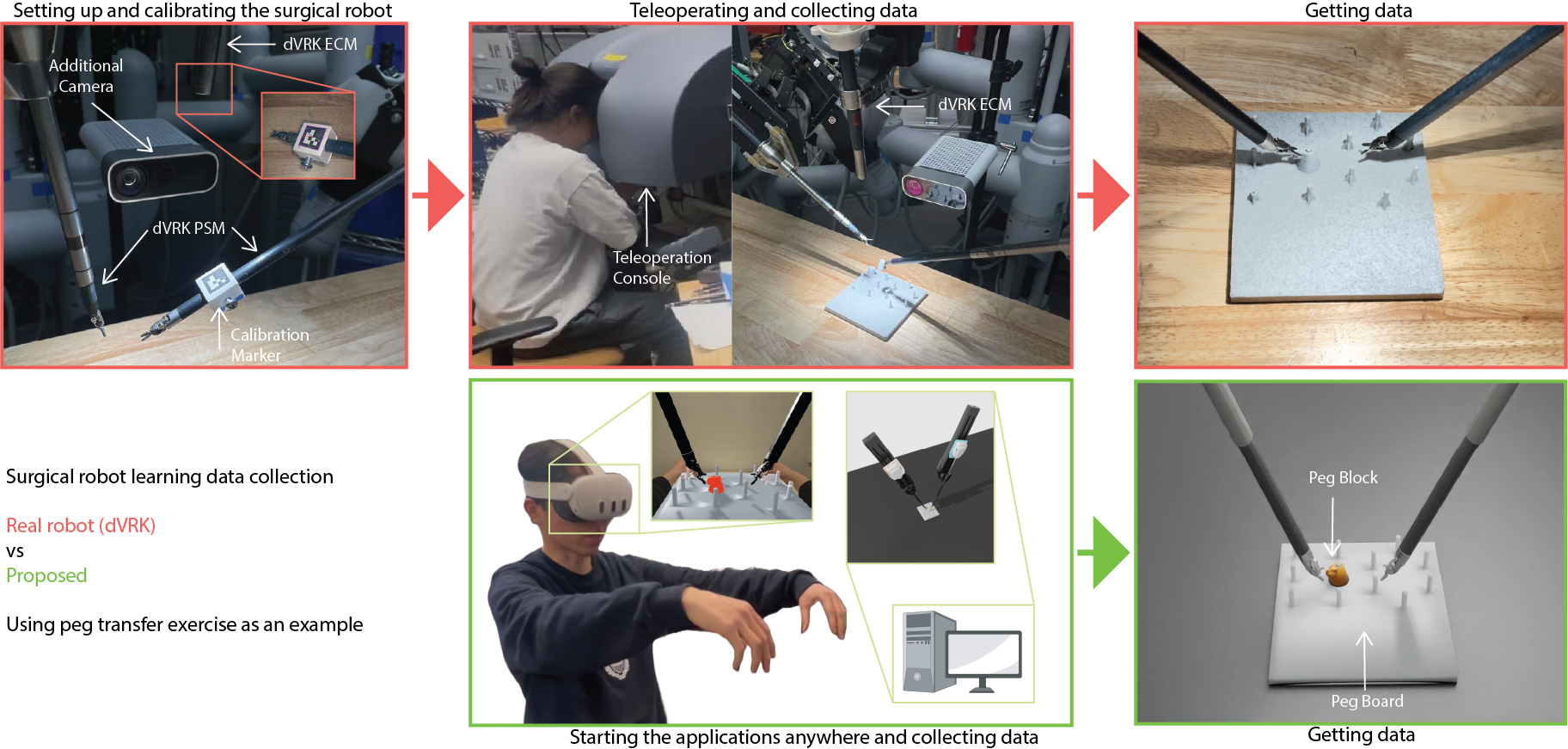}
    \caption{Workflow comparison between using the da Vinci Research Kit (dVRK) \cite{kazanzides2014open} (top) vs the proposed system (bottom) to collect surgical robot learning data. The real robot needs to be set up in the working environment. The preparation includes but is not limited to the calibration processes to obtain the corresponding transformations from the Endoscopic Camera Manipulator (ECM) to the Patient Side Manipulator (PSM). Data collection is then performed by the participants sitting at the teleoperation console. In the proposed approach, the participant only needs a headset to track the hand gestures and a PC to run a high-fidelity simulator.}
    \label{fig:workflow}
\end{figure*}

To address these challenges, researchers have investigated the use of simulation environments \cite{mittal2023orbit, yu2024orbit, wang2023gensim, hua2024gensim2, zhao2023learning} where the training is not bound to real robots. Some works also use mixed reality and novel headset capabilities to enable efficient teleoperation and data collection \cite{mosbach2022accelerating, iyer2024open, cheng2024open, park2024dexhub}. In particular, DART \cite{park2024dexhub} uses AR hand overlay with the robotic manipulators for Internet scale robot data collection, which has shown impressive efficiency improvements on regular primitive tasks. In view of these innovations, we propose \textit{dARt Vinci}, a novel approach that enables scalable data collection and easy access to RMIS teleoperation. Our study is based on the da Vinci surgical robot, a widely used platform in RMIS \cite{kazanzides2014open}, and extensively investigated for teleoperation skill training \cite{ahmidi2017dataset, nagy2019dvrk, long2022integrating, yu2024orbit}. Ten primitive tasks \cite{yu2024orbit} for teleoperation training are used. Our system implements direct communication between an AR headset and a high-fidelity physics engine, enabling more realistic and highly efficient visual feedback. Using an AR interface and a simulator, \textit{dARt Vinci} allows its users to collect data that is intended for robot learning for surgical procedures. By broadening the pool of data contributors, \textit{dARt Vinci} facilitates the creation of large and diverse datasets, which can help the development of robust learning frameworks for surgical robotics. The main contributions of our work are as follows:

\begin{itemize}
    \item An egocentric data collection platform tailored to surgical tasks in imitation or reinforcement learning. The efficiency and usability are validated by a user study.
    \item The system is not bound to the real surgical robot and can be used in any settings where an AR headset and a PC are available so that data collection can be performed at scale.
    \item The design of the text-based data format allows smaller storage instead of massive video data. The produced compact data is re-playable in a high-fidelity physics engine to obtain the video before training.
\end{itemize}

\section{Related Works}

\begin{figure*}[ht]
    \centering
    \includegraphics[width=\textwidth]{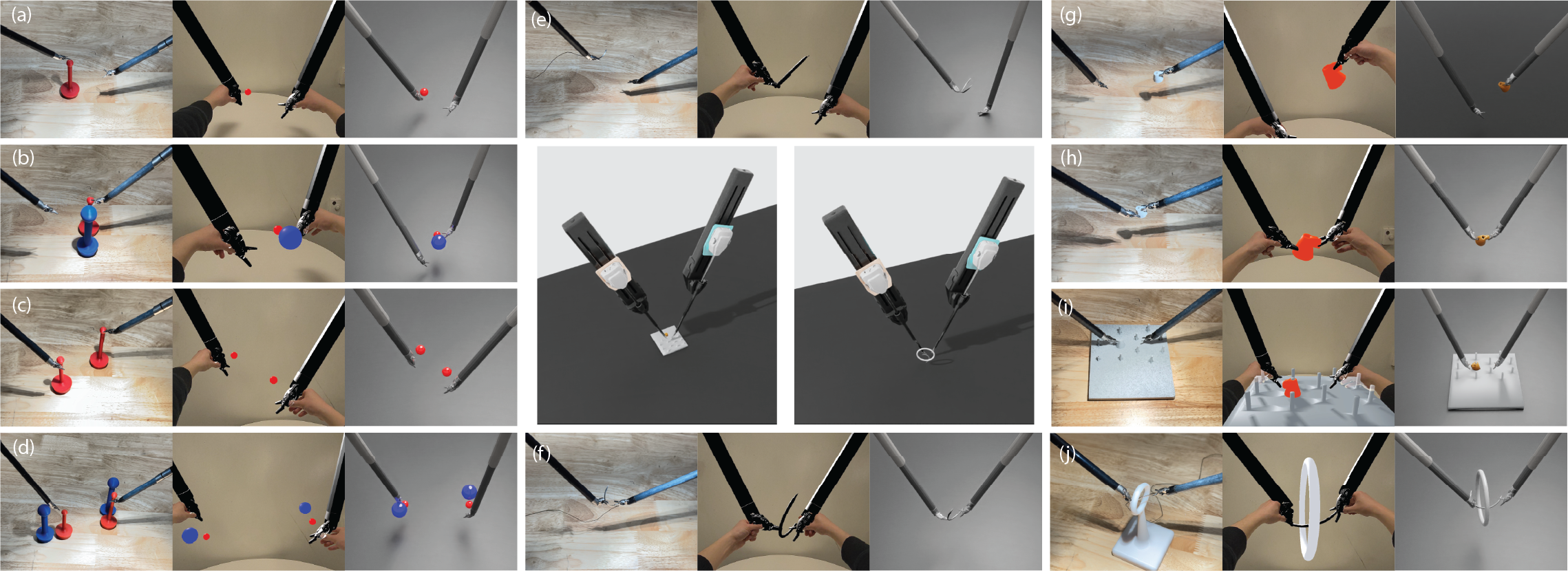}
    \caption{The primitive surgical tasks using dVRK. The center of the Fig. contains the complete view of the dVRK (robot base omitted) while executing peg transfer and needle passing tasks. Panels (a)-(j) demonstrate ten primitive surgical tasks commonly seen in benchmarking surgical robot learning and recognizing surgical gestures \cite{ahmidi2017dataset, hwang2022automating, yu2024orbit}. They are (a) Reach, (b) Reach with Obstacles, (c) Dual Arm Reach, (d) Dual Arm Reach with Obstacles, (e) Suture Needle Lift, (f) Needle Handover, (g) Peg Block Lift, (h) Pick and Transfer, (i) Pick and Place, and (j) Needle Pass Ring. In each panel, from left to right, the views are real robot view, AR headset view while overlaying grippers on hands, and simulator view. }
    \label{fig:demo}
\end{figure*}

\subsection{Robot Learning}

Building on previous imitation learning techniques \cite{pomerleau1988alvinn, brohan2022rt, jang2022bc, dasari2021transformers, florence2022implicit}, more sophisticated algorithmic improvements such as Action Chunk Transformer (ACT) attempt to address compound errors of conventional BC and have shown significant improvement on inaccurate robot platforms \cite{zhao2023learning, kim2024surgical}. Notably, in \cite{kim2024surgical}, Kim \textit{et al.} has found that using egocentric multiview and improved vision are the key to producing reliable results, and our system has an aligned goal. Our system supports the collection of vision data with unconstrained angles of view such that the collection can be tailored to the need.

The stream of research features human demonstrations as learning inputs like in the study by Huang \textit{et al.} \cite{huang2023}, which proposes a demonstration-guided robotic learning paradigm to facilitate training robot policy from human experts. Building these robotic datasets usually involves volunteer participants to collect robotic data teleoperatively \cite{fang2024rh20t, ebert2021bridge, khazatsky2024droid, zhao2023learning, kim2024surgical}. Human-in-the-loop embodied intelligence for surgical robot learning takes human input from haptic devices and maps it to robot state space. Human users then may receive feedback from rendered images. Following this trend, there are consistent reports of robot learning methods for surgical robots. Other notable works include dVRL \cite{richter2019open}, AMBF \cite{varier2022ambf}, SurgicalGYM \cite{schmidgall2024surgical}, and SurRol \cite{xu2021surrol} but do not provide an AR interface for scalable data collection. A work that uses SurRol integrated an AR module \cite{long2022integrating}, but focuses on providing 3D guidance trajectory.

\subsection{Data Collection from Simulation and AR}

The rapid development of robot learning, boosted by the recent wave of embodied AI, necessitates \textit{efficient} and \textit{specialized} data collection for training purposes. As general-purpose robotic learning is being used in specialized applications such as surgical robotics, the criteria of success also become strict, and setting up a real robotic environment and resetting the robot and scenes are no longer trivial efforts.  

A solution to the complicated data collection process is the use of simulators. The behavior cloning work in 1980s \cite{pomerleau1988alvinn} uses a simulated road generator. The relatively low resolution of both the real and simulated vision inputs makes it difficult to distinguish between them. However, capturing more vision details is inevitably important as these algorithms are being deployed in real applications where rich scene content is needed, such as surgical applications. To that end, simulation environments are heavily investigated to make scalable data collection possible. IsaacLab is a unified and modular framework for robot learning that aims to simplify common workflows in robotics research (such as reinforcement learning, learning from demonstrations, and motion planning). It is built on Nvidia IsaacSim to leverage the latest simulation capabilities for photo-realistic scenes and fast and efficient simulation. Recent efforts, including Orbit-Surgical \cite{yu2024orbit}, have leveraged IssacLab's feature to generate robot datasets and benchmark the model's performance. Similar noteworthy foundation frameworks also include the recent platform Genesis \cite{Genesis}.

An inspiring work, DART \cite{park2024dexhub}, allows users to use AR goggles to collect demonstrations under an unlimited number of scenes without having to set up environments physically. Despite the lack of specialized application, the work has successfully demonstrated efficiency by higher data throughput and lower operator fatigue compared to teleoperation using real robots. We build on this concept, applying the design philosophy to surgical robot learning and introducing tailored innovations based on the specific application: dARt Vinci is developed for surgical skills and is made compatible with the existing surgical robot learning frameworks.

\section{Method}

Our system is designed to offer a simple solution for the scalable collection of egocentric surgical robot learning data. By ``simple,'' we mean it is easy to set up and reset, supports remote access, enables replay, and allows for view adjustments without requiring additional hardware or calibration. Our approach combines the augmented rendered overlays and photorealistic physics engine to meet success criteria—scaling data collection without needing a physical robot. Users interact with both virtual and real-world settings through AR goggles, with the Patient Side Manipulators (PSMs) directly overlaid on hand gestures. The comparison of the workflows between using the real robot and our proposed system is shown in Fig. \ref{fig:workflow}. The following introduces the implementation details and the methods used to validate usability.

\subsection{Primitive Tasks}

A complete surgical procedure can be decomposed into numerous primitive tasks. Prior studies have defined and examined these foundational components, as current robotic learning technologies still face challenges with long-horizon actions. In surgical applications, such tasks—including but not limited to the ten illustrated in Fig. \ref{fig:demo}—represent the essential maneuvers used to assess technical skills in laparoscopic surgery training curricula \cite{yu2024orbit}. For instance, peg transfer tasks serve as standardized exercises in minimally invasive procedures, which enhance hand-eye coordination, ambidexterity, and fine motor skills.

\subsection{System Overview}

\begin{figure*}[ht]
    \centering
    \includegraphics[width=\textwidth]{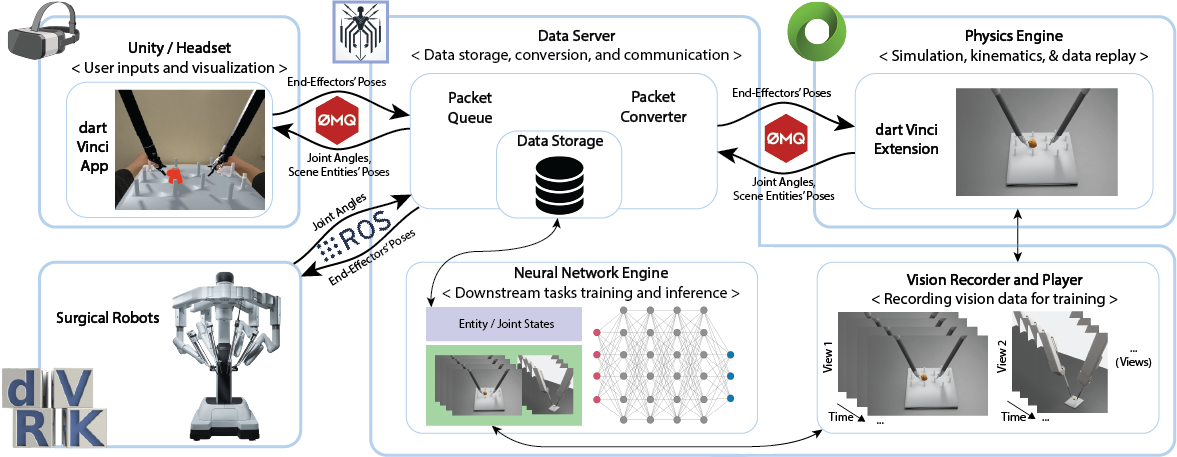}
    \caption{The architecture of the proposed data collection platform. There are three major components in addition to the robot to be used in the experiments: The AR headset is used to visualize the maneuver and track the hand gesture, the data server processes and stores the collected data and passes the end-effector pose to the physics engine, and the physics engine handles the simulation and returns back the relevant states after an update of a simulation frame. The vision player replays the saved data to obtain the video data, and this data is used for robot learning. The real robot can also be connected to the data server for data replaying.}
    \label{fig:architecture}
\end{figure*}

As shown in Fig. \ref{fig:architecture}, the system comprises three main components: the headset application, the data server, and the physics engine. The headset application is built in Unity and deployed on the Meta Quest 3. Each scene contains all the necessary assets for its corresponding tasks, which can be invoked via the physics engine’s user interface. The robot assets are obtained from a previous work by Xu \textit{et al} \cite{xu2021surrol}. The headset application's primary functions are hand gesture tracking and the visualization of scene entities while maneuvering, but not the collection of the vision data, which is done in the high-fidelity simulator (Fig. \ref{fig:architecture}). We employ IsaacSim as our physics engine and have developed a custom extension for streamlined scene management and resetting. In our experiments, the application runs on a laptop equipped with an Intel i9-14900k processor, 32 GB of RAM, and an RTX4090 mobile version GPU.

The data server manages the processing and storage of data packets. These JSON-formatted packets, ranging from 80 to 300 bytes, describe the states of robot joints and scene entities at each sampled timestamp. Each data frame represents the world state at a specific timestamp, with all poses adhering to the right-hand convention to ensure consistency across different simulators. The system operates in real time.

Once the system is running, the user manipulates the target entity with the grippers overlaid on their hands. The hand gesture tracker captures the positions of the finger joints, which serve as the basis for mapping to gripper actions (see Fig. \ref{fig:gripper} and Section \ref{sec:actionmapping}). This process involves data that describes both the pose of the end-effector and the gripper's opening. After a data packet is generated, it is sent to the data server for storage and forwarded to the physics engine, where an inverse kinematics solver converts it into joint angles. In the simulator, each simulation step uses the newly converted joint states to update the scene along with the embedded physics. The updated poses of entities and joint states are then transmitted back to the data server and appended to the historical data. Upon task completion, the data is saved in JSON format. This file captures the entire sequence of robot joint movements and entity poses throughout the task, allowing for replay via our implemented physics player.

\subsection{Action Mapping} \label{sec:actionmapping}

\begin{figure*}[ht]
    \centering
    \includegraphics[width=0.9\textwidth]{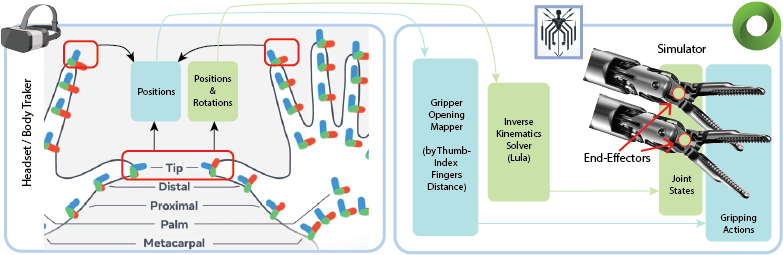}
    \caption{The action mapping from the hand gesture to the PSM manipulator. The system follows the OpenXR Hand Skeleton convention. The distance between the tips of the index finger and the thumb is used to map to the opening and closing of the PSM gripper. The end-effector is anchored on the tip of the thumb whose pose is used in the inverse kinematics solver to derive the joint angles of the PSM arm joints. The yaw joint of the gripper is used as the end of the manipulator.}
    \label{fig:gripper}
\end{figure*}

\begin{figure}
    \centering
    \includegraphics[width=\linewidth]{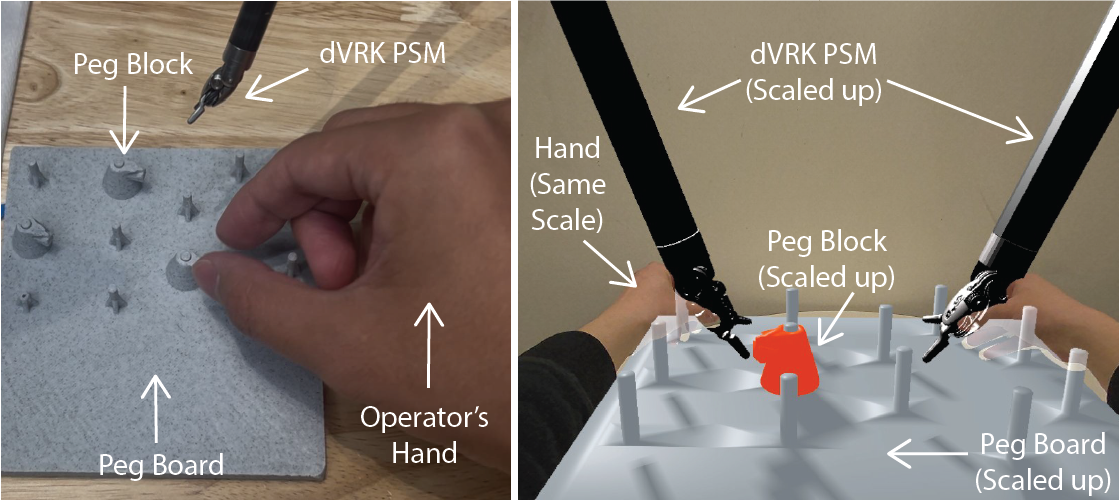}
    \caption{Scaling of the scene entities. The left is a photo to illustrate the actual size of the hand, instruments, and peg board, whereas the right image shows the AR view, where virtual objects are scaled 6x of the true size such that subtle actions can be performed and tolerance for error is increased. }
    \label{fig:scale}
\end{figure}

The hand gesture is mapped to robot actions in real time. To achieve this, the correspondence between the hand and the gripper (end-effector) needs to be established. Here, for dVRK, we define the end-effector center to be the yaw joint of the gripper, highlighted in the simulator panel in Fig. \ref{fig:gripper}. We use OpenXR Hand Skeleton convention for hand gesture capture, where each joint of the hand can be isolated. We use the tip of the thumb with an offset to anchor the end-effector and use the distance between the tip of the index finger and the thumb to be the trigger for the gripping action of the gripper.

The mapping enables intuitive control, where hand position determines the end-effector's movement, and pinching gestures control the gripper's opening and closing. The gripper follows a proportional control scheme: As the distance between the index finger and thumb decreases, the gripper closes; as it increases, the gripper opens. This design closely mimics real-world tool manipulation. 

The physical da Vinci robot provides scaling of the vision, and we also introduce a scaling factor in our approach to obtain finer-grained entity manipulation. The virtual entities in the AR application are scaled up by a factor of six relative to the real world, as shown in Fig. \ref{fig:scale}. This increase in size enables the capture of subtle actions that would otherwise be difficult to perform. If the control and visual representations are at a miniaturized scale, accurate manipulation requires extremely precise hand movements. In contrast, our scaled setup offers a larger tolerance for error, facilitating easier and more precise control. The scaling factor of six was selected empirically during the development but could be changed based on the results of further studies, or could become a parameter adjusted by the user.

\subsection{Physics Engine and Post Data Collection}

IsaacSim is a high-fidelity robotics simulation platform developed by Nvidia. It is designed to help researchers and developers train, test, and deploy robotic applications in a virtual environment. IsaacSim is built on Nvidia’s Omniverse platform, which provides realistic physics simulation, detailed graphics, and simulated sensor data. This makes it an ideal tool for developing complex robotics algorithms before transitioning to real-world hardware. This simulation environment supports integration with common robotics frameworks and middleware. It is ideal for our purpose as the vision data is collected and the fidelity of the captured videos is important. The support of a variety of tools including kinematics solvers helps us for future extension and scaling to additional tasks. It supports customized UI extension which is what we use for our system integration.

Note that the vision data is not collected while the user is running the AR application. Instead, it is recorded by replaying the collected JSON data, which makes getting multi-view data flexible as a virtual camera can be attached anywhere in the simulator and it is not time-sensitive. Nonetheless, recording the vision data while running the AR application is possible, but the system performance can be a bottleneck given the amount of data to be processed in real time. Together with the robot joint states' time series, they are fed into the neural network engine for training.

\subsection{User Study}

The primary objective is to demonstrate the efficiency of data collection compared to using a real robot. We evaluate data collection efficiency in MR versus dVRK by measuring time, data throughput, and usability. The usability is assessed primarily through standardized survey results. For real robot manipulation, we use Patient Side Manipulators (PSMs) with an Endoscopic Camera Manipulator (ECM), after a hand-eye calibration for image data acquisition. The user controls the PSMs directly using the Master Tool Manipulators (MTMs). In contrast, within the headset application, the user manipulates the PSM jaws via hand gestures, with the thumb and index finger performing the grasping motion, as demonstrated in Fig. \ref{fig:demo}.

The user study was approved by the Johns Hopkins Institutional Review Board (HIRB00000701). Each participant completes the same set of tasks using both the headset and the dVRK system, with the order randomized to mitigate learning effects. Every subject is required to perform ten tasks on both platforms. Within a fixed time frame of two minutes per task, participants attempt to complete as many iterations as possible. Fig. \ref{fig:demo} (a–j) illustrates these tasks, displaying screenshots or photographs taken from the real robotic system, the headset, and the simulation platform. For tasks that involve randomness, a random seed ensures that the task resets to a varied starting state. In contrast, when using the real robot, the operator manually resets the task, intentionally repositioning the key object slightly differently to introduce variation. Additional evaluation also involves analyzing the composition of the data relative to previous imitation and reinforcement learning studies.

\begin{table}[h]
    \centering
    \caption{Comparison of data types and sizes for \textit{dARt Vinci} and real robot.}
    \label{table:table1}
    \begin{tabular}{@{}l|c|c@{}}
        \toprule
        \textbf{Data Type} & \textbf{\textit{dARt Vinci}} & \textbf{Real Robot} \\ 
        \midrule
        PSM end-effector poses & \checkmark & \checkmark \\ 
        PSM joint states & \checkmark & \checkmark \\ 
        ECM stereo images & (replayable) & \checkmark \\ 
        All entity poses & \checkmark & --- \\ 
        \midrule
        \textbf{Frequency} & 72 Hz & 30 Hz\\
        \midrule
        \textbf{Total Size} & 1.0 GB & 436.0 GB \\
        \bottomrule
    \end{tabular}
\end{table}

\section{Results and Discussion}

\begin{figure*}
    \centering
    \includegraphics[width=0.9\textwidth]{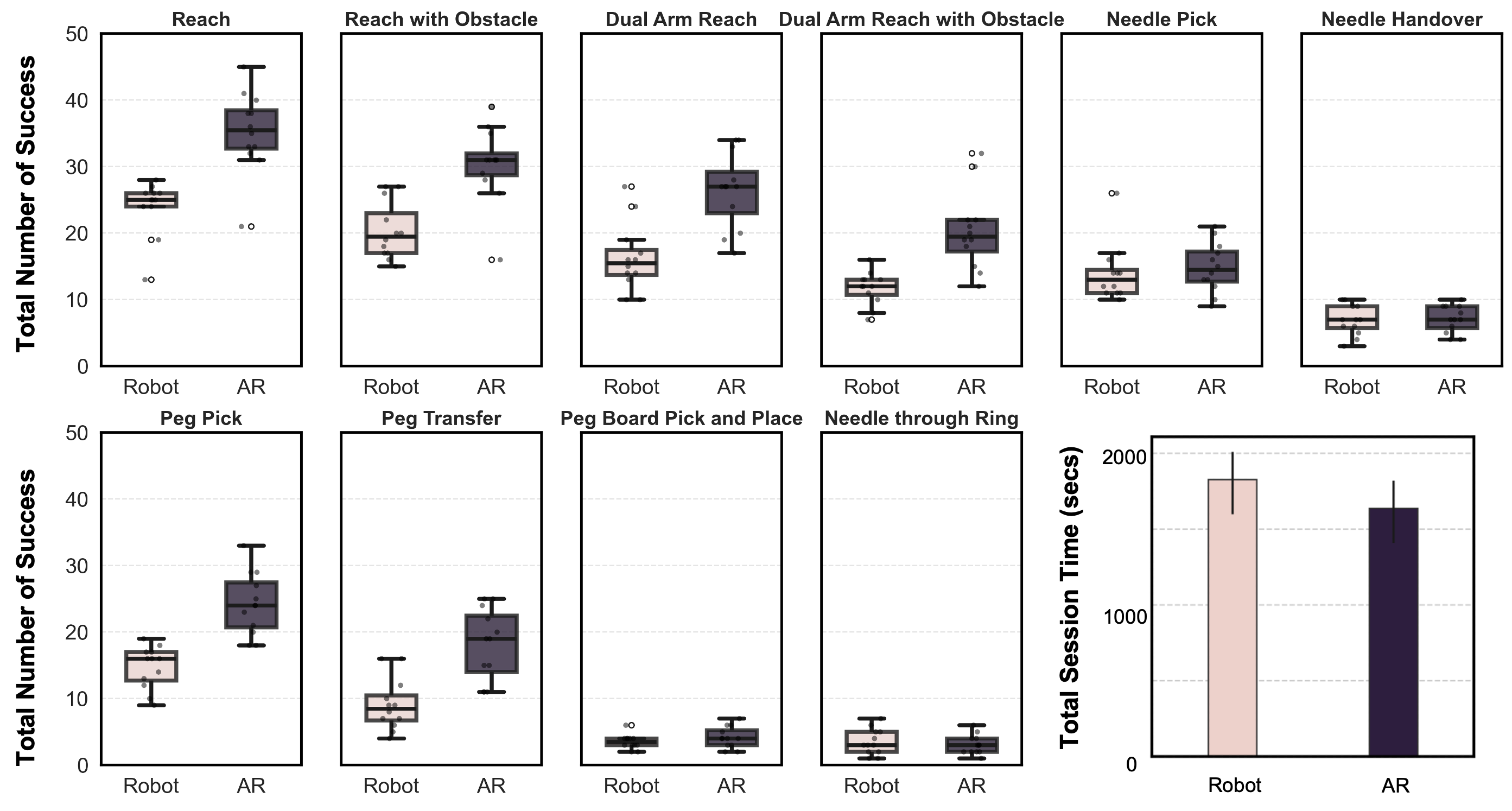}
    \caption{Data throughput comparison between \textit{dARt Vinci} and real dVRK system.  Participants were asked to perform the tasks as many times as possible for 2 minutes. The total times of success are reported for each task. The lower right panel compares the total session time between the two systems. Total time includes the time for switching setups between tasks but does not include the time to set up and calibrate the real robot.}
    \label{fig:data_throughput}
\end{figure*}

We recruited a total of 12 subjects, consisting of six female and six male participants. These subjects are all novice users without trained surgical skills. Each subject was instructed to perform ten distinct tasks on both the robotic platform (dVRK) and our proposed platform. For each task, participants were asked to repeat the operation as many times as possible within a 2-minute time frame. The total number of successful attempts and completions are reported in Fig.~\ref{fig:data_throughput} and Table~\ref{table:success_rates}. On average, the proposed method resulted in a higher number of attempts and successful completions.  

On average, the data throughput is improved by 41\%. The paired t-test found that the majority of the tasks are significantly different (with $p<0.05$). \textit{dARt Vinci} not only enabled a higher number of successful attempts but also maintained the overall success rate, as shown in Table \ref{table:success_rates}. For simpler tasks, such as reaching a designated target, a successful attempt was defined as the moment when the jaw of the instrument made contact with the target sphere(s) without any form of collision with the obstacle sphere(s). Among these tasks, \textit{\textit{dARt Vinci}} significantly increased the number of successful attempts by reducing the time required to reset the task scenario, thereby improving data throughput. For more complex tasks involving peg manipulation, a successful attempt was defined as the peg being securely held by the instrument's jaw throughout the entire procedure and at the moment of task completion. Any unintended peg drop was considered a failure. 

\begin{figure}
    \centering
    \includegraphics[width=\linewidth]{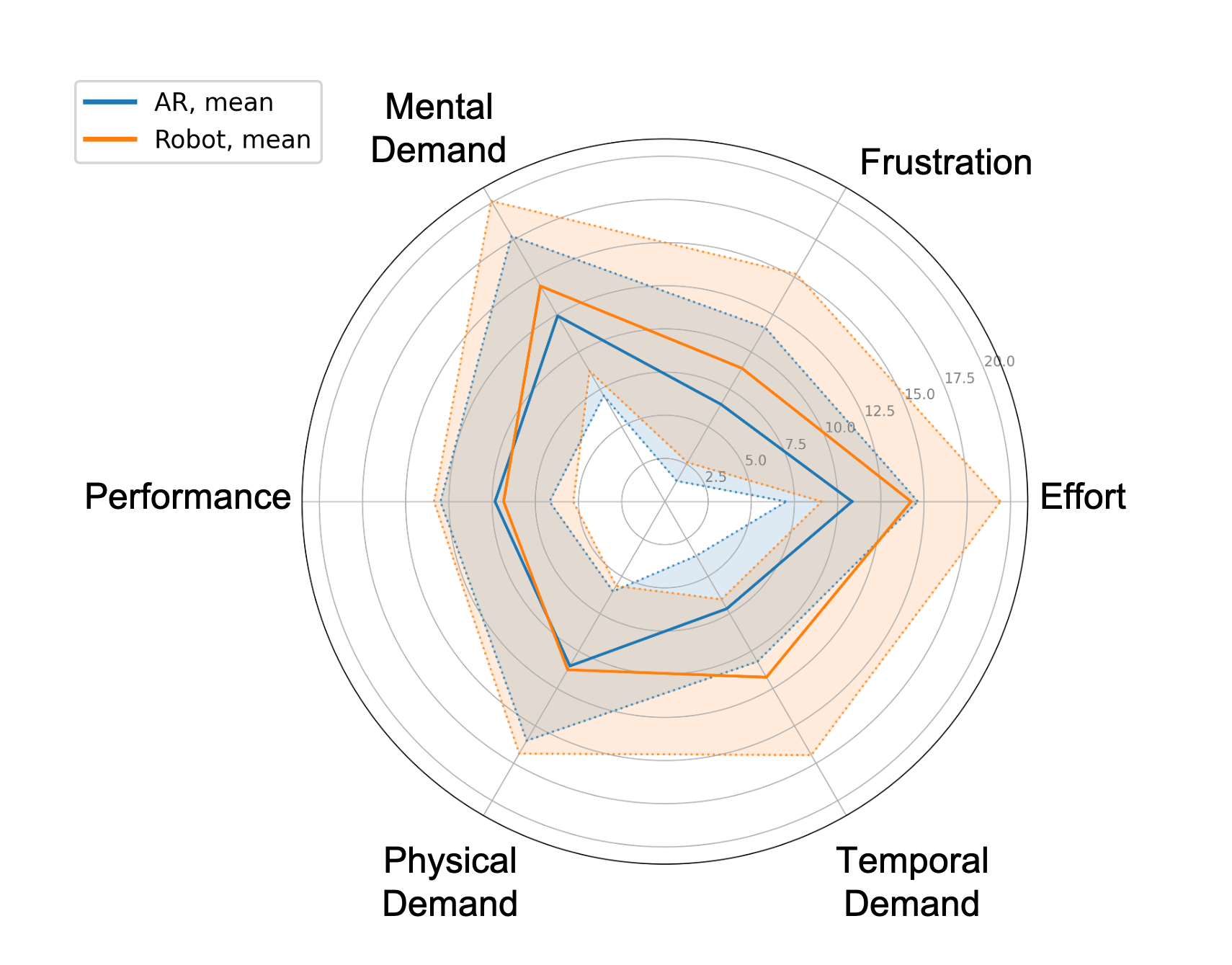}
    \caption{Radar plot displaying the mean and standard deviation for each dimension of the NASA Task Load Index (TLX) across all participants under two different conditions. The dotted lines, along with the shaded areas, represent the standard deviation for each condition and dimension. Values closer to the center signify a lower task load.}
    \label{fig:tlx}
\end{figure}

The NASA TLX results in Fig. \ref{fig:tlx} further highlight the benefits of using \textit{dARt Vinci}. While most workload dimensions did not show statistically significant differences between the two platforms ($p > 0.05$), a notable exception is \textit{Temporal Demand}, where participants reported significantly lower workload when using \textit{dARt Vinci} ($p = 0.0189$). Additionally, although not statistically significant, trends in the data suggest that \textit{dARt Vinci} generally resulted in a lower perceived workload across most dimensions compared to the real robot.

However, no significant improvement was observed in needle-related tasks, likely due to the challenges associated with fine collider modeling and the precise handling of model penetration, given the needle's small size. Nevertheless, \textit{dARt Vinci} facilitated a more efficient data collection process. The average total session time for dVRK was 1822.85 seconds, not even including the time required to set up and calibrate the real robot, whereas for \textit{dARt Vinci}, it was reduced to 1642.40 seconds, with $p \leq 0.05$. This statistically significant reduction indicates improved efficiency in data collection on the proposed platform. Table \ref{table:table1} also shows that the data storage needed for our approach is much lower than the real robot approach (1:436) even though ours is at a higher frequency (double), primarily because the video data can be obtained by replaying the text-based data post-data-collection.

Although the advantages are clear, some limitations and possible solutions are worth noting. The user study is performed on a relatively small cohort of subjects. We may validate the usability with a larger group and divide the expert and novice participants to obtain more subtle insights on how they may approach the maneuvers differently. The physics engine community is also experiencing rapid advancement in this wave of revolutionary embodied AI deployment. The fidelity of the simulators is being improved and it is especially meaningful to us in tasks that involve cutting and piercing deformable materials. It is also important to benchmark the data collected from our egocentric platform in the existing imitation and reinforcement learning algorithms, which can be the focus of a follow-up study.

\begin{table}[t]
    \centering
    \caption{Comparison of Success Rates Between Robot and AR}
    \label{table:success_rates}
    \renewcommand{\arraystretch}{1.2} 
    \begin{tabular}{l|cc}
        \toprule
        \textbf{Task} & \textbf{Robot Success \%} & \textbf{AR Success \%} \\
        \midrule
        Reach & 100 & 100 \\
        Reach w/ Obstacle & 100 & 100 \\
        Dual Arm Reach & 100 & 100 \\
        Dual Arm Reach w/ Obstacle & 100 & 98 \\
        Needle Pick & 98 & 88 \\
        Needle Handover & 78 & 66 \\
        Peg Pick & 98 & 98 \\
        Peg Transfer & 91 & 95 \\
        Peg Board Pick & 79 & 78 \\
        Needle through Ring & 55 & 56 \\
        \bottomrule
    \end{tabular}
        
\end{table}

\section{Conclusion}

In this work, we propose a framework to collect egocentric robot learning data without having to set up a real surgical robot. It allows scalable data collection and resetting of the scenes in one click. Our user study has demonstrated the efficiency of the data collection and the usability of the system. The data throughput is improved across all tasks by 41\% on average compared to the real robot settings and the total time for a set of experiments can be reduced by 10\% on average. The collected data is compact, which saves data storage by more than 400 times even with a doubled sampling frequency.

\bibliographystyle{IEEEtran}
\bibliography{bib} 

\end{document}